\documentclass[10pt,twocolumn,letterpaper]{article}

\usepackage{iccv}
\usepackage{times}
\usepackage{epsfig}
\usepackage{graphicx}
\usepackage{amsmath}
\usepackage{amssymb}
\usepackage{color}

\usepackage[breaklinks=true,bookmarks=false]{hyperref}

\iccvfinalcopy 

\def\iccvPaperID{****} 

\begin{document}
\pagestyle{empty}
\title{Multi-label Image Recognition by Recurrently Discovering Attentional Regions}

\author{Zhouxia Wang$^{1,2}$ \and Tianshui Chen$^{1}$ \and Guanbin Li$^{1,3}$ \and Ruijia Xu$^{1}$ \and Liang Lin$^{1,2,3}$ \thanks{ Zhouxia Wang and Tianshui Chen contribute equally to this work and share first-authorship. Corresponding author is Liang Lin (Email: linliang@ieee.org). This work was supported by the State Key Development Program under Grant 2016YFB1001004,  Special Program for Applied Research on Super Computation of the NSFC-Guangdong Joint Fund (second phase), and CCF-Tencent Open Research Fund (NO.AGR20160115).} \\
\and \small $^1$ Sun Yat-sen University, China \and \small $^2$ SenseTime Group Limited\\
\and \small $^3$ Engineering Research Center for Advanced Computing Engineering Software of Ministry of Education, China \\
{\tt\small \{{wzhoux, chtiansh, xurj3}\}@mail2.sysu.edu.cn, liguanbin@mail.sysu.edu.cn,  linliang@ieee.org }
}

\maketitle

\begin{abstract}
This paper proposes a novel deep architecture to address multi-label image recognition, a fundamental and practical task towards general visual understanding. Current solutions for this task usually rely on an extra step of extracting hypothesis regions (i.e., region proposals), resulting in redundant computation and sub-optimal performance. In this work, we achieve the interpretable and contextualized multi-label image classification by developing a recurrent memorized-attention module. This module consists of two alternately performed components: i) a spatial transformer layer to locate attentional regions from the convolutional feature maps in a region-proposal-free way and ii) an LSTM (Long-Short Term Memory) sub-network to sequentially predict semantic labeling scores on the located regions while capturing the global dependencies of these regions. The LSTM also output the parameters for computing the spatial transformer. On large-scale benchmarks of multi-label image classification (e.g., MS-COCO and PASCAL VOC 07), our approach demonstrates superior performances over other existing state-of-the-arts in both accuracy and efficiency.
\end{abstract}

\section{Introduction}
Recognizing multiple labels of images is a fundamental yet practical problem in computer vision, as real-world images always contain rich and diverse semantic information. Besides the challenges shared with single-label image classification (e.g., large intra-class variation caused by viewpoint, scale, occlusion, illumination), multi-label image classification is much more difficult since accurately predicting the presence of multiple object categories usually needs understanding the image in depth (e.g., associating semantic labels with regions and capturing their dependencies).

\begin{figure}[!t]
   \centering
   \includegraphics[width=1.0\linewidth]{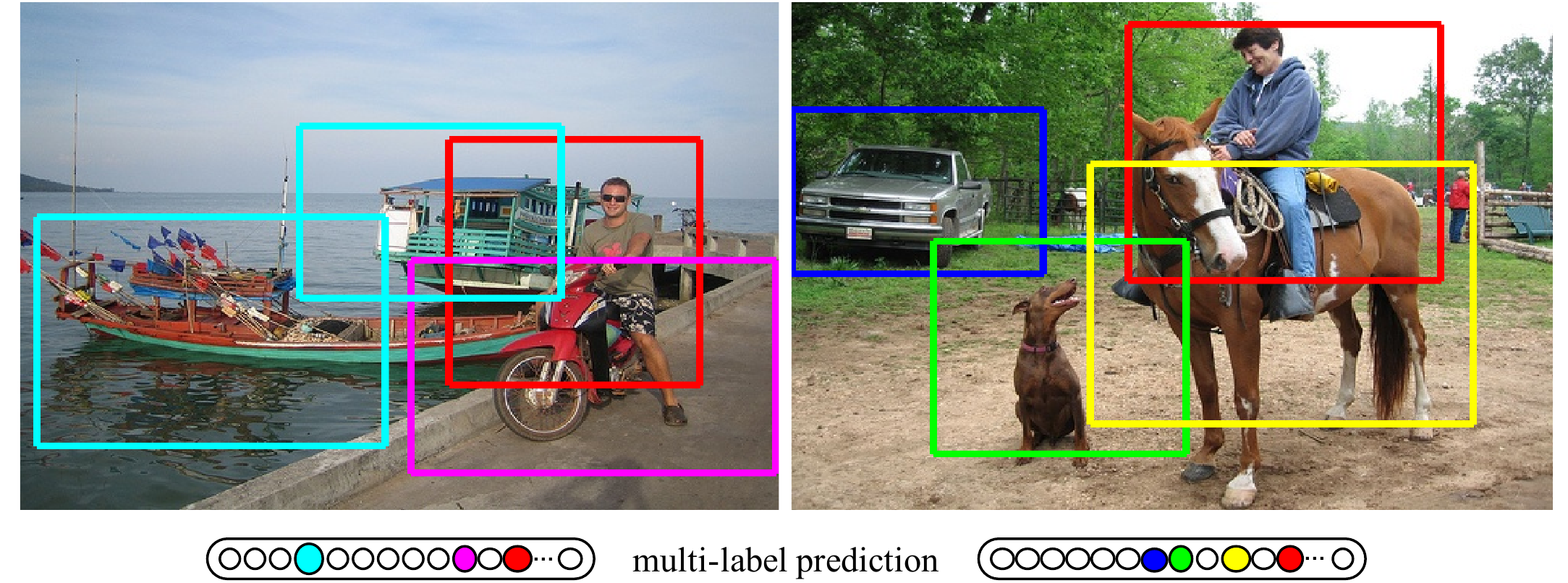}
   \caption{Multi-label image recognition with discovered attentional regions by our approach. These regions (highlighted by different colors) corresponding to the semantic labels (visualized below the images) are contextualized and discriminative in terms of classification, although they may not preserve object boundaries well.}
   \label{fig:vis_attention_region}
\end{figure}

Recently, convolutional neural networks (CNNs) \cite{krizhevsky2012imagenet, simonyan2014very} achieve great success in visual recognition/classification tasks by learning powerful feature representations from raw images, and they have been also applied to the problem of multi-label image classification by combining with some object localization techniques \cite{yang2016exploit, wei2015hcp}. The resulting common pipeline usually involves two steps. A batch of hypothesis regions are first produced by either exploiting bottom-up image cues \cite{uijlings2013selective} or casting extra detectors \cite{cheng2014bing}, and these regions are assumed to contain all possible foreground objects in the image. A classifier or neural network is then trained to predict the label score on these hypothesis regions, and these predictions are aggregated to achieve the multi-label classification results. Despite acknowledged successes, these methods take the redundant computational cost of extracting region proposals and usually over-simplify the contextual dependencies among foreground objects, leading to a sub-optimal performance in complex scenarios. Recently, Wang et al. \cite{wang2016cnn} proposed to jointly characterize the semantic label dependency and the image-label relevance by combining recurrent neural networks (RNNs) with CNNs. However, their model disregards the explicit associations between semantic labels and image contents, and lacks fully exploiting the spatial context in images. In contrast to all these mentioned methods, we introduce an end-to-end trainable framework that explicitly discovers attentional regions over image scales corresponding to multiple semantic labels and captures the contextual dependencies of these regions from a global perspective. No extra step of extracting hypothesis regions is needed in our approach. Two examples generated by our approach are illustrated in Figure \ref{fig:vis_attention_region}.

To search for meaningful and discriminative regions in terms of multi-label classification, we propose a novel recurrent memorized-attention module, which is combined with convolutional neural networks in our framework. Specifically, this module consists of two components: i) a spatial transformer layer to locate attentional regions on the convolutional maps and ii) an LSTM (Long-Short Term Memory) sub-network to sequentially predict the labeling scores over the attentional regions and output the parameters of the spatial transformer layer. Notably, the global contextual dependencies among the attentional regions are naturally captured (i.e., memorized) together with the LSTM sequential encoding. And the two components are alternately performed during the recurrent learning. In this way, our approach enables to learn a contextualized and interpretable region-label relevance while improving the discriminability for multi-label classification.

The main contributions of this work are three-fold. 
{\flushleft $\bullet$} We develop a proposal-free pipeline for multi-label image recognition, which is capable of automatically discovering semantic-aware regions over image scales and simultaneously capturing their long-range contextual dependencies.

{\flushleft $\bullet$} We further propose three novel constraints on the spatial transformer, which help to learn more meaningful and interpretable regions, and in turn, facilitate multi-label classification.

{\flushleft $\bullet$} We conduct extensive experiments and evaluations on large-scale benchmarks such as PASCAL VOC \cite{everingham2010pascal} and Microsoft COCO \cite{lin2014microsoft}, and demonstrate the superiority of our proposed model in both recognition accuracy and efficiency over other leading multi-label image classification methods.

\section{Related Works}
The performance of image classification has recently witnessed a rapid progress due to the establishment of large-scale labeled datasets (i.e., PASCAL VOC~\cite{everingham2010pascal}, COCO~\cite{lin2014microsoft}) and the fast development of deep CNNs~\cite{simonyan2014very,he2015deep}. In recent years, many researchers have attempted to adapt the deep CNNs to multi-label image recognition problem and have achieved great success.

\subsection{Multi-label image recognition}
Traditional multi-label image recognition methods apply the bag-of-words (BOW) model to solve this problem~\cite{chen2012hierarchical, dong2013subcategory}. Although performing well on the simple benchmarks, these methods may fail in classifying images with complex scenes since BOW based models depend largely on the hand-crafted low-level features. In contrast, features learned by deep models have been confirmed to be highly versatile and far more effective than the hand-crafted features. Since this paper focuses on deep learning based multi-label image recognition, we discuss the relevant works in the following context.

Recently, there have been attempts to apply deep learning to multi-label image recognition task~\cite{sharif2014cnn,simonyan2014very,yang2016exploit, wei2015hcp,wang2016cnn}. Razavian et al.~\cite{sharif2014cnn} applies off-the-shelf features extracted from deep network pretrained on ImageNet ~\cite{russakovsky2015imagenet} for multi-label image classification. Gong et al. ~\cite{gong2013deep} propose to combine convolutional architectures with an approximate top-k ranking objective function for annotating multi-label images. Instead of extracting off-the-shelf deep features, Chatfield et al.~\cite{chatfield2014return} fine tune the network with the target multi-label datasets, which can learn task-specific features and thus boost the classification performance. To better consider the correlations between labels instead of treating each label independently, traditional graphical models are widely incorporated, such as Conditional Random Field~\cite{ghamrawi2005collective}, Dependency Network~\cite{guo2011multi}, and co-occurrence matrix~\cite{xue2011correlative}. Recently, Wang et al.~\cite{wang2016cnn} utilize the RNNs to learn a joint image-label embedding to characterize the semantic label dependency as well as the image-label relevance.

\begin{figure*}[htbp]
   \centering
   \includegraphics[width=0.90\linewidth]{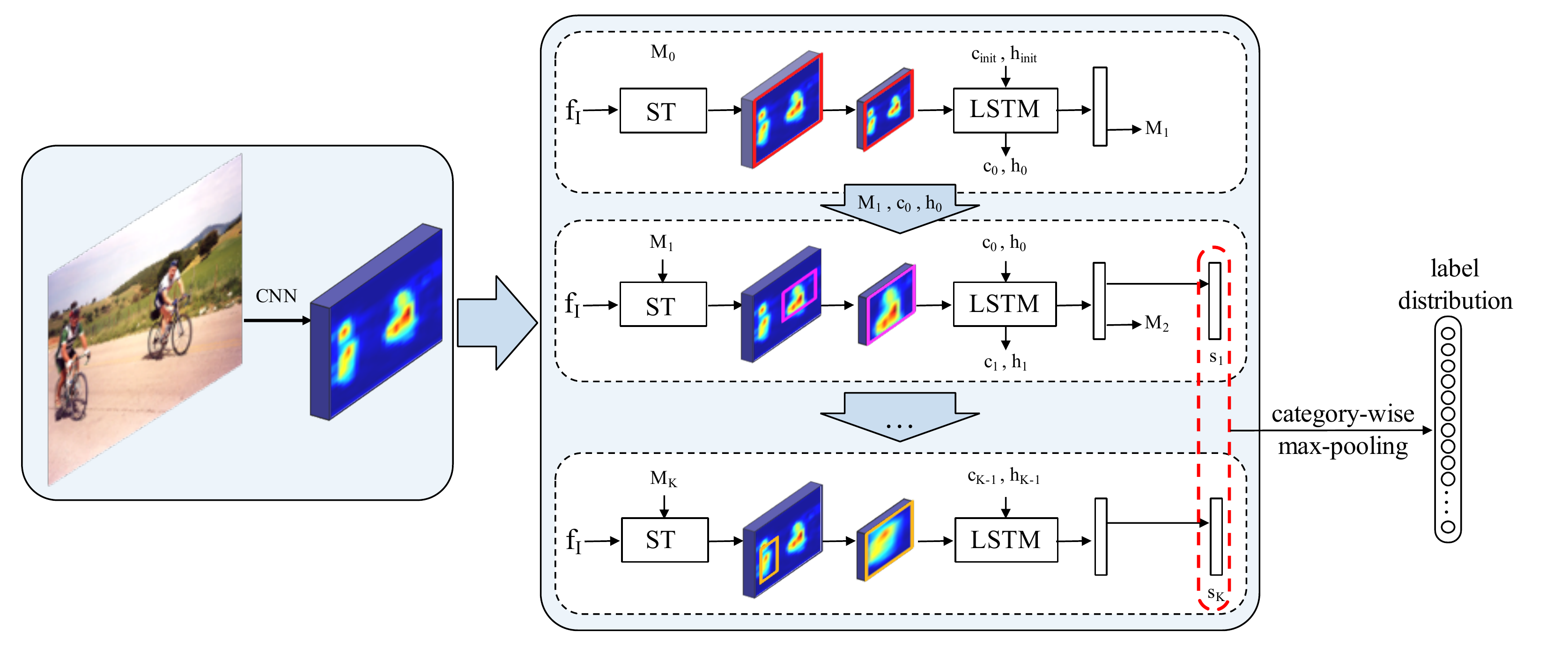}
   \caption{Overview of our proposed framework for multi-label image recognition. Our model iteratively locates the attentional regions corresponding to semantic labels and predicts the score for the current region. }
   \label{fig:framework}
\end{figure*}

All of the aforementioned methods consider extracting the features of the whole image with no spatial information, which on one hand were unable to explicitly perceive the corresponding image regions to the detected classification labels, and on the other hand, were extremely vulnerable to the complex background. To overcome this issue, some researchers propose to exploit object proposals to only focus on the informative regions, which effectively eliminate the influences of the non-object areas and thus demonstrate significant improvement in multi-label image recognition task~\cite{yang2016exploit, wei2015hcp}. More specifically, Wei et al.~\cite{wei2015hcp} propose a Hypotheses-CNN-Pooling framework to aggregate the label scores of each specific object hypotheses to achieve the final multi-label predictions. Yang et al.~\cite{yang2016exploit} formulate the multi-label image recognition problem as a multi-class multi-instance learning problem to incorporate local information and enhance the discriminative ability of the features by encoding the label view information. However, these object proposals based methods are generally not efficient with the preprocessing step of object proposal generation being the bottleneck. Moreover, the training stage is not perfect and can hardly be modeled as an end-to-end scheme in both training and testing. In this paper, we propose to incorporate a recurrent memorized-attention module in the neural network to simultaneously locate the attentional regions and predict the labels on various located regions. Our proposed method does not resort to the extraction of object proposals and is thus very efficient and can be trained in an end-to-end mode.

\subsection{Visual attention model}
Attention model has been recently applied to various computer vision tasks, including image classification ~\cite{mnih2014recurrent, ba2014multiple}, saliency detection ~\cite{kuen2016recurrent}, and image captioning ~\cite{xu2015show}. Most of these works use the recurrent neural network for sequential attentions, and optimized their models with reinforcement learning technique. Works \cite{mnih2014recurrent, ba2014multiple} formulate a recurrent attention model and apply it to the digital classification tasks for which the images are low-resolution with a clean background, using the small attention network. The model is non-differential and addressed with reinforcement learning to learn task-specific policies. Jaderberg et al. \cite{jaderberg2015spatial} propose a differential spatial transformer module which could be used to extract attentional regions with any spatial transformation, including scaling, rotation, transition, and cropping. Moreover, it could be easily integrated into the neural network and optimized using the standard back-propagation algorithm without reinforcement learning.

\section{Model}
Figure \ref{fig:framework} illustrates the architecture of the proposed model. The input image $I$ is first fed into a VGG-16 ConvNet without additional object proposals. The network first processes the whole image with several convolutional (conv) and max pooling layers to produce the conv feature maps, denoted as $\mathbf{f}_I$. Here, we use the conv feature maps from the last conv layer (i.e., conv5\_3). The recurrent memorized-attention module, comprising a spatial transformer (ST)~\cite{jaderberg2015spatial} and an LSTM network ~\cite{hochreiter1997long} that work collaboratively in an iterative manner, predicts the label distributions directly from the input image features. Specifically, in one iterative procedure, the ST locates an attentional region for the LSTM, and the LSTM predicts the scores regarding this region for multi-label classification and simultaneously updates the parameters of ST. Finally, the scores from several attentional regions are fused to achieve the final label distribution.

\subsection{ST for Attentional Region Localization}
We briefly introduce the spatial transformer (ST)~\cite{jaderberg2015spatial} for completeness before diving deep into the recurrent memorized-attention module. ST is a sample-based differential module that spatially transforms its input maps to the output maps with a given size which correspond to a sub-region of the input maps. It is convenient to embed an ST layer in the neural network and train it with the standard back-propagation algorithm. In our model, the ST is incorporated in the recurrent memorized-attention module for the localization of attentional regions.

\begin{figure}[!t]
   \centering
   \includegraphics[width=0.9\linewidth]{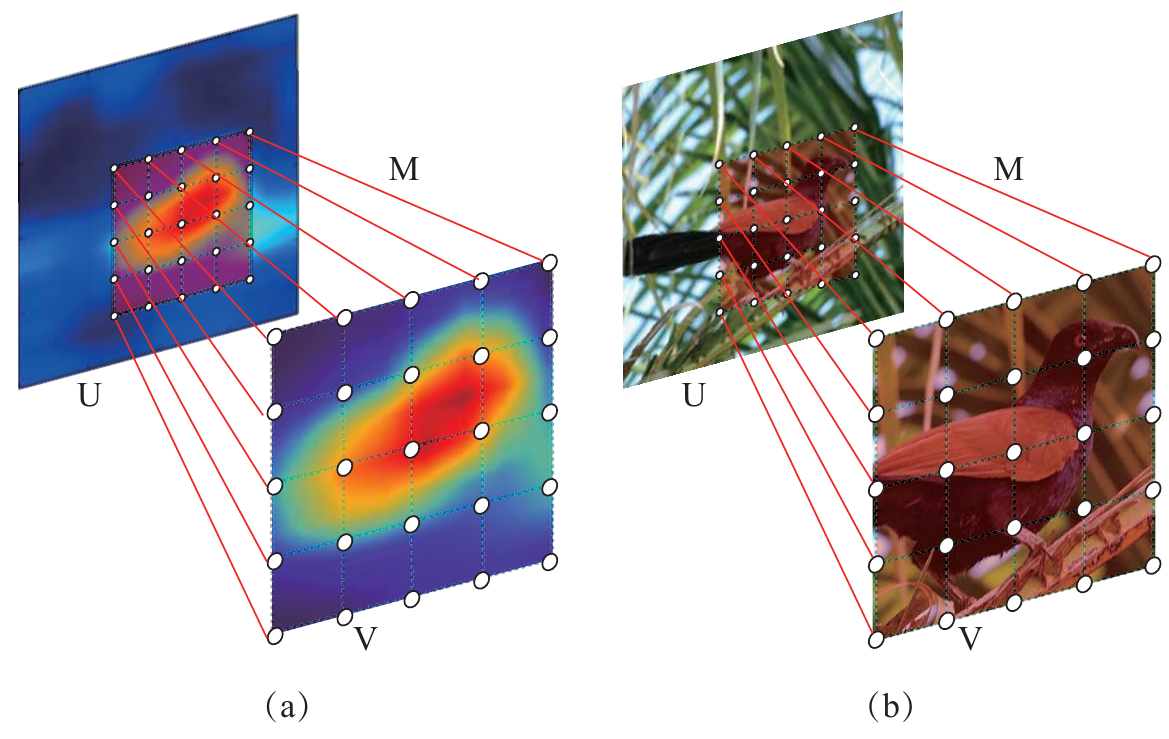}
   \caption{Illustration of coordinate grid mapping on (a) the feature maps and (b) the corresponding input image.}
   \label{fig:st_map}
\end{figure}

Formally, the ST layer extracts features of an attentional region, denoted as $\mathbf{f}_k$, from the feature maps $\mathbf{f}_I$ of the whole input image. The computational procedure is as follows. A transformation matrix $\mathbf{M}$ is first estimated by a localization network (explained later). After that, the corresponding coordinate grid in $\mathbf{f}_I$ is obtained, based on the coordinates of $\mathbf{f}_k$. Then the sampled feature maps $\mathbf{f}_k$ that correspond to the attentional region are generated by bilinear interpolation. Fig. \ref{fig:st_map} shows an example of coordinate mapping. As we aim to locate the attentional regions, we constrain the transformation matrix $\mathbf{M}$ to involve only cropping, translation and scaling, expressed as

\begin{equation}
      \mathbf{M}=   \left[\begin{matrix}
         s_x & 0 & t_x \\
         0 & s_y & t_y \\
      \end{matrix}\right] \\,
\end{equation}
where $s_x$, $s_y$, $t_x$, $t_y$ are the scaling and translation parameters. In our model, we apply a standard neural network to estimate these parameters to facilitate an end-to-end learning scheme.

\subsection{Recurrent Memorized-Attention Module}
The core of our proposed model is the recurrent memorized-attention module, which combines the recurrent computation process of an LSTM network and a spatial transformer. It iteratively searches the most discriminative regions, and predicts the scores of label distribution for them. In this subsection, we introduce this module in detail.

In the $k$-th iteration, our model searches an attentional region, and extracts the corresponding features by applying the following spatial transformer, expressed as

\begin{equation}
   \mathbf{f}_k=\mathrm{st}(\mathbf{f}_I, \mathbf{M}_k),
   \mathbf{M}_k=   \left[\begin{matrix} s_x^k & 0 & t_x^k \\ 0 & s_y^k & t_y^k \\\end{matrix}\right] \\,
\end{equation}
where $\mathrm{st}(\cdot)$ is the spatial transformation function, and $\mathbf{M}_k$ is the transformation matrix estimated in the previous round by the localization network. We initialize the attentional region with the whole image at the first iteration, i.e., the initial transformation matrix is set to be
 \begin{equation}
      \mathbf{M}_0=   \left[\begin{matrix}
         1 & 0 & 0 \\
         0 & 1 & 0 \\
      \end{matrix}\right] \\.
      \label{eqn:initial_matrix}
\end{equation}
Note that we apply the spatial transformer operation on the feature maps $\mathbf{f}_I$ instead of the input image to avoid repeating the computational intensive convolutional processes. The LSTM takes the sampled feature map $\mathbf{f}_k$ as input to compute the memory cell and hidden state. The computation process can be expressed as

\begin{equation}
   \begin{split}
      \mathbf{x}_k &= \mathrm{relu}(  \mathbf{W}_{fx}\mathbf{f}_k + \mathbf{b}_x) \\
      \mathbf{i}_k &= \sigma(\mathbf{W}_{xi}\mathbf{x}_k + \mathbf{W}_{hi}\mathbf{h}_{k-1} +   \mathbf{b}_i) \\
      \mathbf{g}_k &= \sigma(\mathbf{W}_{xg}\mathbf{x}_k + \mathbf{W}_{hg}\mathbf{h}_{k-1} + \mathbf{b}_g) \\
      \mathbf{o}_k &= \sigma(\mathbf{W}_{xo}\mathbf{x}_k + \mathbf{W}_{ho}\mathbf{h}_{k-1} + \mathbf{b}_o) \\
      \mathbf{m}_k &= \tanh(\mathbf{W}_{xm}\mathbf{x}_k + \mathbf{W}_{hm}\mathbf{h}_{k-1} + \mathbf{b}_m) \\
      \mathbf{c}_k &= \mathbf{g}_k \odot \mathbf{c}_{k-1} + \mathbf{i}_k \odot \mathbf{m}_k \\
      \mathbf{h}_k &= \mathbf{o}_k \odot \mathbf{c}_k
   \end{split}
   \label{eqn:LSTM}
\end{equation}
where $\mathrm{relu}(\cdot)$ is the rectified linear function, $\sigma(\cdot)$ is the sigmoid function, $\tanh(\cdot)$ is the hyperbolic tangent function; $\mathbf{h}_{k-1}$ and $\mathbf{c}_{k-1}$ are the hidden state and memory cell of previous iteration; $\mathbf{i}_k$, $\mathbf{g}_k$,  $\mathbf{o}_k$ and $\mathbf{m}_k$ are the outputs of the input gate, forget gate, output gate, and input modulation gate, respectively. These multiplicative gates can ensure the robust training of LSTMs as they work well in exploding and vanishing gradients \cite{hochreiter1997long}.

The memory cell $\mathbf{c}_k$ encodes the useful information of previous $(k-1)$ regions, and it is possible to benefit our task in the following two aspects. First, previous works ~\cite{xue2011correlative, wang2016cnn} have shown that different categories of objects exhibit strong co-occurrence dependencies. Therefore, it helps to recognize objects within the current attentional region aided by ``remembering" information of previous ones. Second, it is expected that our model can find out all relevant and useful regions for classification. Simultaneously considering the information of previous regions is a feasible approach that implicitly enhances the diversity and complementarity among the attentional regions.

\noindent\textbf{Update rule of $\mathbf{M}$}. Given the hidden state $\mathbf{h}_k$,  the classifier and localization network can be expressed as

\begin{equation}
   \begin{split}
         \mathbf{z}_k &= \mathrm{relu}(\mathbf{W}_{hz}\mathbf{h}_k + \mathbf{b}_z) \\
	\mathbf{s}_k &=\mathbf{W}_{zs}\mathbf{z}_k+\mathbf{b}_s, k\neq0 \\
	\mathbf{M}_{k+1} &= \mathbf{W}_{zm}\mathbf{z}_k + \mathbf{b}_m
   \end{split}
\end{equation}
where $\mathbf{s}_k$ is the predicted score distribution of the $k$-th region, and $\mathbf{M}_{k+1}$ is the transformation matrix for the next iteration. Note that at the first iteration ($k=0$), we make no prediction of $\mathbf{s}$ and just estimate the matrix $\mathbf{M}$ because no attentional region is obtained initially.

\noindent\textbf{Category-wise max-pooling}.
The iterations are repeated for $K+1$ times, resulting in $K$ score vectors $\{\mathbf{s}_1,\mathbf{s}_2,\dots,\mathbf{s}_K\}$, where $\mathbf{s}_k=\{s_k^1,s_k^2,\dots,s_k^C\}$ denotes the scores over $C$ class labels. Following ~\cite{wei2015hcp}, we employ the category-wise max-pooling to fuse the scores into the final result $\mathbf{s}=\{s^1,s^2,\dots,s^C\}$. It simply maximizes out the scores over regions for each category
 \begin{equation}
      s^c=\mathrm{max}(s_{1}^c,s_{2}^c,\dots,s_{K}^c), c=1,2,\dots,C.
\end{equation}

\section{Learning}

\subsection{Loss for Classification}
We employ the Euclidean loss as the objective function following \cite{wei2015hcp, yang2016exploit}. Suppose there are $N$ training samples, and each sample $x_i$ has its label vector $\mathbf{y}_i=\{y_{i}^{1},y_{i}^{2},\dots,y_{i}^{C}\}$. $y_{i}^c$ $(c=1,2,\dots,C)$ is assigned as 1 if the sample is annotated with the class label $c$, and 0 otherwise. The ground-truth probability vector of the $i$-th sample is defined as $\mathbf{\hat{p}}_i=\mathbf{y}_i/||\mathbf{y}_i||_1$.
Given the predicted probability vector $\mathbf{p}_i$
 \begin{equation}
      p_{i}^c= \frac{\exp(s_{i}^c)}{\sum_{c'=1}^{C}\exp(s_{i}^{c'})} \ c=1,2,\dots,C,
\end{equation}
and the classification loss function is expressed as
 \begin{equation}
      \mathcal{L}_{\textrm{cls}}=\frac{1}{N}\sum_{i=1}^N\sum_{c=1}^C(p_{i}^c-\hat{p}_{i}^c)^2.
\end{equation}

\subsection{Loss for Attentional Region Constraints}
As discussed above, we obtain the final result by aggregating the scores of the attentional regions. Thus, we hope that the attentional regions selected by our model contain all of the objects in the input image. If one object is left out unexpectedly, an inevitable error occurs because the LSTM network has never seen this object during the prediction procedure. We experimentally found that the proposed model can be trained with the defined classification loss, however, has notable drawbacks:
\begin{itemize}
  \item \textbf{Redundancy}. The ST layer usually picks up the same region that corresponds to the most salient objects. As a result, it would be difficult to retrieve all of the objects appearing in the input image, since the set of attentional regions are redundant.
  \item \textbf{Neglect of tiny objects}. The ST layer tends to locate regions in a relatively large size and ignores the tiny objects, which hampers the classification performance.
  \item \textbf{Spatial flipping.} The selected attentional region may be mirrored vertically or horizontally.
\end{itemize}
To address these issues, we further define a loss function that consists of three constraints on the parameters of the transformation matrix $\mathbf{M}$.

\begin{figure}[!t]
   \centering
   \includegraphics[width=0.9\linewidth]{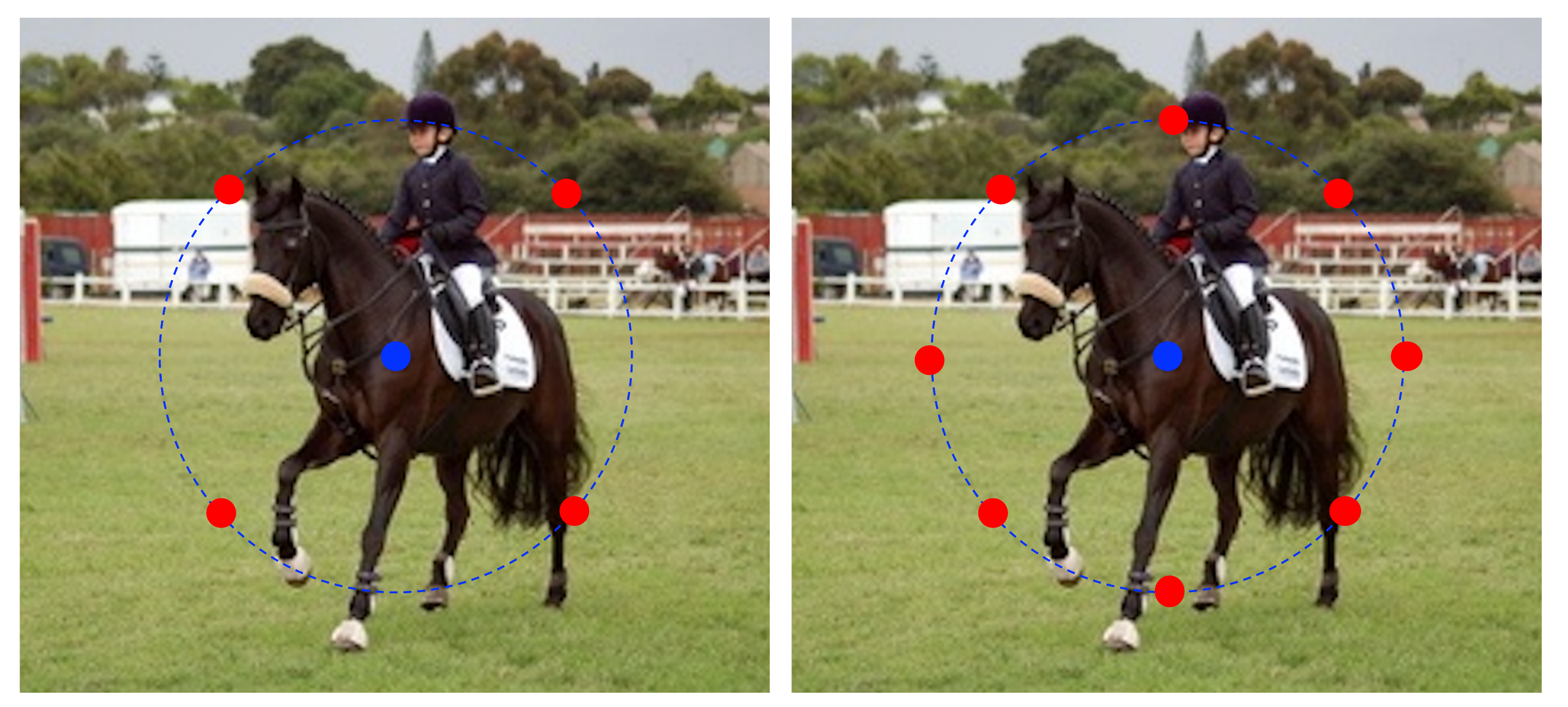}
   \caption{Anchor selection for left: $K$=5 and right: $K$=9.}
   \label{fig:anchor}
\end{figure}

\noindent\textbf{Anchor constraint}. It would be better if the attentional regions scatter over different semantic regions in the image. For the first iteration, adding no constraint helps to find the most discriminative region. After that, we push the other $({K-1})$ attentional regions away from the image center by an anchor constraint. We draw a circle of radius $\frac{\sqrt{2}}{2}$ centered on the image center, and pick up the anchor points on the circle uniformly, as depicted in Figure \ref{fig:anchor}. We use $K=5$ in the experiments, so four anchor points are generated at $(0.5,0.5)$, $(0.5,-0.5)$, $(-0.5,0.5)$, and $(-0.5,-0.5)$, respectively\footnote{The range of coordinate is rescaled to [-1, 1]}. The anchor constraint is formulated as
\begin{equation}
   \ell_{A}=\frac{1}{2}\{(t_x^k-c_x^k)^2+(t_y^k-c_y^k)^2\},
\end{equation}
where $(c_x^k, c_y^k)$ is the location of the $k$-th anchor point.

\noindent\textbf{Scale constraint}. This constraint attempts to push the scale parameters in a certain range, so that the located attentional region will not be too large in size. It can be formulated as
\begin{equation}
   \ell_{S}= \ell_{s_x}+ \ell_{s_y},
\end{equation}
in which
\begin{equation}
\begin{split}
 \ell_{s_x}=(\max (|s_x| - \alpha, 0))^2 \\
  \ell_{s_y}=(\max (|s_y| - \alpha, 0))^2
\end{split}
\end{equation}
where $\alpha$ is a threshold value, and it is set as 0.5 in our experiments.

\noindent\textbf{Positive constraint}. The last one also constrains the scale parameters. Positive constraint prefers a transformation matrix with positive scale parameters, leading to attentional regions that are not be mirrored:
\begin{equation}
    \ell_{P}=\max(0, \beta-s_x)+\max(0, \beta-s_y),
\end{equation}
where $\beta$ is a threshold value, set as 0.1 in our experiments.

Finally, we combine the aforementioned three types of constraints on the parameters of the transformation matrix to define a loss of localization of attentional regions. It is formulated as the weighted sum of the three components:
\begin{equation}
   \mathcal{L}_{\textrm{loc}}=  \ell_{S}+\lambda_1 \ell_{A}+\lambda_2 \ell_{P},
\end{equation}
where $\lambda_1$ and $\lambda_2$ are the weighted parameters, and they are set as 0.01 and 0.1, respectively.

Our model is jointly trained with the classification loss and the localization loss, so the overall loss function can be expressed as
 \begin{equation}
      \mathcal{L}=\mathcal{L}_{\textrm{cls}}+\gamma\mathcal{L}_{\textrm{loc}}.
\end{equation}
We set the balance parameter $\gamma$ as 0.1 since the classification task is dominated in our model. Optimization is performed using the recently proposed Adam algorithm \cite{kingma2014adam} and standard back-propagation.

\section{Experiments}
\subsection{Settings}
\noindent\textbf{Implementation details.} We implemented our method on the basis of Caffe~\cite{jia2014caffe} for deep network training and testing. In the training stage, we employed a two-step training mechanism to initialize the convolutional neural network following \cite{wei2015hcp}. The CNN is first pre-trained on the ImageNet, a large scale single label classification dataset, and further fine-tuned on the target multi-label classification dataset. The learned parameters are used to initialize the parameters of the corresponding layers in our proposed model, while the parameters of other newly added layers in our network are initialized with Xavier algorithm~\cite{glorot2010understanding,he2015delving} rather than manual tuning.  All the training images are first resized to $N\times{N}$, and randomly cropped with a size of $(N-64)\times(N-64)$. The training samples are also augmented by horizontal flipping. In our experiments, we trained two models with $N=512$ and $N=640$, respectively. Both of the models are optimized using Adam with a batch size of 16, momentum of 0.9 and 0.999. The learning rate is set to 0.00001 initially and divided by 10 after 30 epochs. We trained the models for about 45 epochs for each scale, and selected the model with the lowest validation loss as the best model for testing.

In the testing phase, we follow~\cite{krizhevsky2012imagenet} to perform ten-view evaluation across different scales. Specifically, we first resized the input image to $N\times{N}$ ($N=512, 640$), and extracted five patches (i.e., the four corner patches and the center patch) with a size of $(N-64)\times(N-64)$, as well as their horizontally flipped versions. Instead of repeatedly extracting features for each patch, the model feeds the N$\times$N image to the VGG-16 ConvNet, and crops the features on the conv5\_3 features maps accordingly to achieve the features of all patches. Then, for each patch, the model extracts the features for each located attentional region, and eventually aggregates the features of this patch by max-pooling. The image representation is obtained via averaging the features of all the patches. At last, we trained a one-vs-rest SVM classifier for each category using the LIBLINEAR library~\cite{fan2008liblinear}. We test our model on a single NVIDIA GeForce GTX TITAN-X, and it takes about 150ms for ten-view evaluation for scale 512, and about 200 ms for scale 640. It reduces the execution time by more than an order of magnitude, compared with previous proposal-based methods, e.g., HCP~\cite{wei2015hcp}, which costs about 10s per image.

\noindent\textbf{Evaluation metrics.} We use the same evaluation metrics as \cite{wang2016cnn}. For each image, we assign top $k$ highest-ranked labels to the image, and compare with the ground-truth labels. We compute the overall precision, recall, F1 (OP, OR, OF1) and per-class precision, recall, F1 (CP, CR, CF1) in Eq. \ref{eqn:metric}. Following \cite{wei2015hcp,dong2013subcategory}, we also apply average precision (AP) for each category, and the mean average precision (mAP) over all categories as well. Generally, overall F1, per-class F1, and mAP are relatively important.

\begin{equation}
   \begin{split}
         OP&=\frac{\sum_{i}N_{i}^{c}}{\sum_{i}N_{i}^{p}}\\
         OR&=\frac{\sum_{i}N_{i}^{c}}{\sum_{i}N_{i}^{g}}\\
         OF1&=\frac{2 \times OP \times OR}{OP+OR}
   \end{split}
      \begin{split}
         CP&=\frac{1}{C}\sum_{i}\frac{N_{i}^{c}}{N_{i}^{p}}\\
         CR&=\frac{1}{C}\sum_{i}\frac{N_{i}^{c}}{N_{i}^{g}}\\
         CF1&=\frac{2 \times CP \times CR}{CP+CR},
   \end{split}
   \label{eqn:metric}
\end{equation}
where $C$ is the number of labels, $N_{i}^{c}$ is the number of images that are correctly predicted for the $i$-th label, $N_{i}^{p}$ is the number of predicted images for the $i$-th label, $N_{i}^{g}$ is the number of ground truth images for the $i$-th label.

\subsection{Comparison with State-of-the-art Methods}
To validate the effectiveness of our model, we conduct the experiments on two benchmarks, PASCAL VOC 2007 ~\cite{everingham2010pascal} and Microsoft COCO~\cite{lin2014microsoft}. VOC 2007 is the most widely used benchmark, and most works have reported the results on this dataset. We compare the performance of our proposed method against the following state-of-the-art approaches:  FeV+LV-20-VD~\cite{yang2016exploit}, HCP~\cite{wei2015hcp}, RLSD \cite{zhang2016multi}, CNN-RNN~\cite{wang2016cnn}, VeryDeep~\cite{simonyan2014very} and CNN-SVM ~\cite{sharif2014cnn} on the VOC 2007 dataset. MS-COCO is released later than VOC and more challenging. Recent works have also used this benchmark for evaluation. We compare with CNN-RNN~\cite{wang2016cnn}, RLSD \cite{zhang2016multi} and WARP~\cite{gong2013deep} on the COCO dataset as well.

\begin{table*}[htp]
\centering
\scriptsize
\begin{tabular} 
{p{1.6cm}|p{0.3cm}p{0.3cm}p{0.3cm}p{0.3cm}p{0.4cm}p{0.3cm}p{0.20cm}p{0.20cm}p{0.3cm}p{0.3cm}p{0.3cm}p{0.3cm}p{0.35cm}p{0.4cm}p{0.45cm}p{0.3cm}p{0.35cm}p{0.3cm}p{0.3cm}p{0.5cm}|p{0.5cm}}
\hline
\centering Methods  & aero & bike & bird & boat & bottle & bus & car & cat & chair & cow & table & dog & horse & mbike & person & plant & sheep & sofa & train & tv & mAP \\
\hline
\hline
\centering CNN-SVM~\cite{sharif2014cnn} & 88.5 & 81.0  & 83.5 &  82.0 & 42.0 & 72.5 & 85.3 & 81.6 & 59.9 & 58.5 & 66.5 & 77.8 & 81.8 & 78.8 & 90.2 & 54.8 & 71.1 & 62.6 & 87.2  & 71.8 & 73.9 \\
\centering CNN-RNN~\cite{wang2016cnn} & 96.7 & 83.1 & 94.2 & 92.8 & 61.2 & 82.1 & 89.1 & 94.2 & 64.2 & 83.6 & 70.0 & 92.4 & 91.7 & 84.2 & 93.7 & 59.8 & 93.2 & 75.3 & \textcolor[rgb]{1,0,0}{99.7} & 78.6 & 84.0 \\
\centering VeryDeep~\cite{simonyan2014very}  & \textcolor[rgb]{1,0,0}{98.9} & 95.0 & \textcolor[rgb]{0,0,1}{96.8} & 95.4 & 69.7 & 90.4 & 93.5 & 96.0 & 74.2 & 86.6 & \textcolor[rgb]{1,0,0}{87.8} & 96.0 & 96.3 & 93.1 & 97.2 & 70.0 & 92.1 & 80.3 & 98.1 & 87.0 & 89.7\\
\centering RLSD \cite{zhang2016multi} & 96.4 &  92.7 & 93.8 & 94.1 & 71.2 &  92.5 &  94.2 & 95.7 & 74.3 &  90.0 & 74.2  & 95.4 & 96.2 &  92.1 & 97.9 & 66.9 & 93.5 & 73.7 & 97.5 & 87.6 & 88.5\\
\centering HCP~\cite{wei2015hcp}  & \textcolor[rgb]{0,0,1}{98.6} & 97.1  & \textcolor[rgb]{1,0,0}{98.0} & 95.6 & \textcolor[rgb]{1,0,0}{75.3}&{\noindent\color{red}{94.7}}  & \textcolor[rgb]{0,0,1}{95.8} &\textcolor[rgb]{1,0,0}{97.3} & 73.1 & 90.2 & 80.0 & \textcolor[rgb]{1,0,0}{97.3} & 96.1 & \textcolor[rgb]{0,0,1}{94.9} & 96.3 & 78.3 & \textcolor[rgb]{1,0,0}{94.7} & 76.2 & 97.9 & \textcolor[rgb]{1,0,0}{91.5} & 90.9\\
\centering FeV+LV~\cite{yang2016exploit}  & 97.9 & 97.0 & 96.6 & 94.6 & 73.6 & \textcolor[rgb]{0,0,1}{93.9} & \textcolor[rgb]{1,0,0}{96.5}& 95.5 & 73.7 & 90.3 & 82.8 & 95.4 & \textcolor[rgb]{1,0,0}{97.7} & \textcolor[rgb]{1,0,0}{95.9} & \textcolor[rgb]{1,0,0}{98.6} & 77.6 & 88.7 & 78.0 & 98.3 & 89.0 & 90.6\\
\hline
\centering Ours (512)  & 98.5 & 96.7 & 95.6 & 95.7 & 73.7 & 92.1 & \textcolor[rgb]{0,0,1}{95.8} & 96.8 &  \textcolor[rgb]{1,0,0}{76.5} & \textcolor[rgb]{1,0,0}{92.9}& 87.2 & 96.6 & \textcolor[rgb]{0,0,1}{97.5} & 92.8 & 98.3 & 76.9 & 91.3 & \textcolor[rgb]{1,0,0}{83.6} & \textcolor[rgb]{0,0,1}{98.6} & 88.1 & \textcolor[rgb]{0,0,1}{91.3} \\
\centering Ours (640)  & 97.7 & \textcolor[rgb]{0,0,1}{97.3} & 96.4 & \textcolor[rgb]{0,0,1}{95.8} & 74.6 & 91.9 & \textcolor[rgb]{1,0,0}{96.5} & 96.7 & \textcolor[rgb]{0,0,1}{75.2} & 89.9 & 87.1 & 96.0 & 96.9 & 93.2 & 98.4 & \textcolor[rgb]{0,0,1}{81.3} & 93.4 & 81.3 & 98.3 & 88.5 & \textcolor[rgb]{0,0,1}{91.3} \\
\centering Ours  & \textcolor[rgb]{0,0,1}{98.6} & \textcolor[rgb]{1,0,0}{97.4} & 96.3 & \textcolor[rgb]{1,0,0}{96.2} & \textcolor[rgb]{0,0,1}{75.2} & 92.4 & \textcolor[rgb]{1,0,0}{96.5} & \textcolor[rgb]{0,0,1}{97.1} & \textcolor[rgb]{1,0,0}{76.5} & \textcolor[rgb]{0,0,1}{92.0} & \textcolor[rgb]{0,0,1}{87.7} & \textcolor[rgb]{0,0,1}{96.8} & \textcolor[rgb]{0,0,1}{97.5}  & 93.8 & \textcolor[rgb]{0,0,1}{98.5} & \textcolor[rgb]{1,0,0}{81.6}& \textcolor[rgb]{0,0,1}{93.7} & \textcolor[rgb]{0,0,1}{82.8} & \textcolor[rgb]{0,0,1}{98.6} &\textcolor[rgb]{0,0,1}{89.3} & \textcolor[rgb]{1,0,0}{91.9} \\
\hline
\end{tabular}
\vspace{1pt}
\caption{ Comparison of AP and mAP in \% of our model and state-of-the-art methods on the PASCAL VOC 2007 dataset. The best results and second best results are highlighted in {\color{red}{red}} and {\color{blue}{blue}}, respectively. Best viewed in color.}
\label{table:comparision_voc07}
\end{table*}

\subsubsection{Performance on the VOC 2007 dataset}
The PASCAL VOC 2007 dataset contains 9,963 images from 20 object categories, which is divided into train, val and test sets. We train our model on the trainval set, and evaluate the performance on the test set, following other competitors. Table \ref{table:comparision_voc07} presents the experimental results. The previous best-performing methods are HCP and FeV+LV, which achieve a mAP of 90.9\% and 90.6\%, respectively. Both of them share a similar two-step pipeline: they first extract the object proposals of the image, and then aggregate the features of them for multi-label classification. Different from them, our method is proposal-free since the attentional regions are selected by the ST layer that works collaboratively with the LSTM network. In this way, the interaction between attentional region localization and classification is well explored, leading to improvement in performance. Our proposed method achieves a mAP of 91.9\%, that outperforms previous state-of-the-art algorithms. Note that our model learned with a single scale of 512 or 640 also surpasses previous works. This better demonstrates the effectiveness of the proposed method.

\begin{table}[htp]
\centering
\scriptsize
\begin{tabular}{c|ccc|ccc}
\hline
\centering Methods  & C-P & C-R & C-F1 & O-P & O-R & O-F1  \\
\hline
\hline
\centering  WARP~\cite{gong2013deep}  & 59.3 & 52.5 & 55.7 & 59.8 & 61.4 & 60.7   \\
\centering  CNN-RNN~\cite{wang2016cnn}  & 66.0 & 55.6 & 60.4 & 69.2 & \textcolor[rgb]{1,0,0}{66.4} & 67.8 \\
\centering RLSD \cite{zhang2016multi}  & 67.6 &  57.2 &  62.0 & 70.1 & \textcolor[rgb]{0,0,1}{63.4} & 66.5 \\
\hline
\centering  Ours (512)  & 77.7 &  \textcolor[rgb]{0,0,1}{58.1} &  \textcolor[rgb]{0,0,1}{66.5} & 83.4 & 62.3 & 71.3 \\
\centering  Ours (640)  & \textcolor[rgb]{0,0,1}{78.0} & 57.7 & 66.3 &  \textcolor[rgb]{0,0,1}{83.8} & 62.3 &  \textcolor[rgb]{0,0,1}{71.4} \\
\centering  Ours & \textcolor[rgb]{1,0,0}{79.1} &\textcolor[rgb]{1,0,0}{58.7} & \textcolor[rgb]{1,0,0}{67.4} & \textcolor[rgb]{1,0,0}{84.0} &  63.0 & \textcolor[rgb]{1,0,0}{72.0} \\
\hline
\end{tabular}
\vspace{4pt}
\caption{Comparison of our model and state-of-the-art methods on the MS-COCO dataset. The best results and second best results are highlighted in {\color{red}{red}} and {\color{blue}{blue}}, respectively. Best viewed in color.}
\label{table:coco_comparison}
\end{table}
\subsubsection{Performance on the MS-COCO dataset}
The MS-COCO dataset is primarily built for object detection, and it is also widely used for multi-label recognition recently. It comprises a training set of 82,081 images, and a validation set of 40,137 images. The dataset covers 80 common object categories, with about 3.5 object labels per image. The label number for each image also varies considerably, rendering MS-COCO even more challenging. As the ground truth labels of the test set are not available, we evaluate the performance of all the methods on the validation set instead. We follow~\cite{wang2016cnn} to select the top $k=3$ labels for each image, and filter out the labels with probabilities lower than a threshold 0.5, so the label number of some images would be less than 3.

We compare the overall precision, recall, F1, and per-class precision, recall, F1 in Table \ref{table:coco_comparison}. Our model outperforms the existing methods by a sizable margin. Specifically, it achieves a per-class F1 score of 67.4\% and an overall F1 score of 72.0, improving those of the previously best method by 5.4\% and 4.2\%, respectively. Similar to the results on VOC, the model learned with a single scale also beats the state-of-the-art approaches.

\subsection{Ablation Study}
In this subsection, we perform ablative studies to carefully analyze the contribution of the critical components of our proposed model.

\subsubsection{Attentional regions v.s. object proposals}
\label{section:attention_region}
One of the main contributions of this work is that our model is capable of discovering the discriminative regions, which facilitates the task of multi-label image classification compared with proposal-based methods. In this subsection, we present a comparison to reveal the fact that attentional regions have significant advantages against object proposals.

\begin{figure}[!t]
   \centering
   \includegraphics[width=1.0\linewidth]{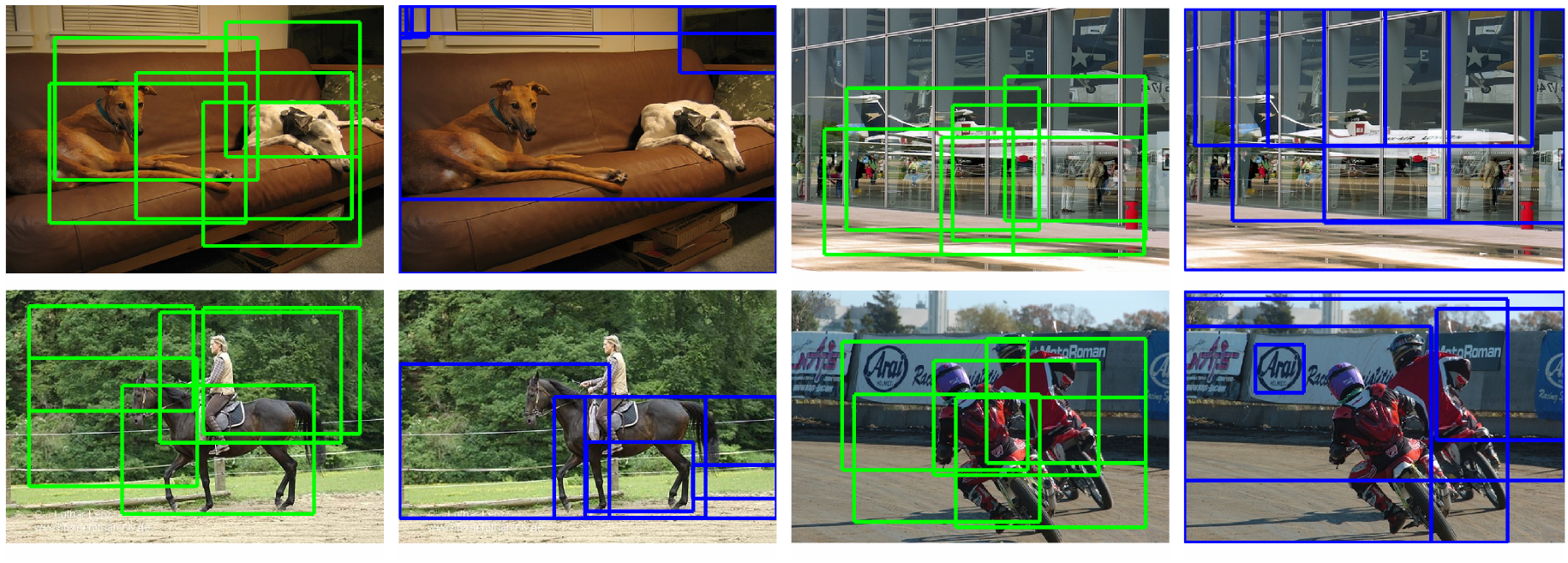}
   \caption{Comparison of visualization of the attentional regions (indicated by green boxes) located by our method, and the object proposals (indicated by blue boxes) generated by EdgeBox. }
   \label{fig:attention_region}
\end{figure}

Proposal-based methods are proved to be powerful for objectness detection. However, satisfactory recall rates are difficult to achieve until thousands of proposals are provided. In addition, it is extremely time-consuming to examine all of the provided proposals with a deep network. As an example, although HCP selects some representative proposals, it still needs 500 proposals to obtain desired performance. Besides, computing the object proposals also introduces additional computational overhead. In contrast, our model utilizes an efficient spatial transformation layer to find out a small number of discriminative regions, making the model runs much faster. Here we also present the visualization results of attentional regions discovered by our model, and those generated by EdgeBox ~\cite{zitnick2014edge}, a representative proposal method. For our method, $K$ is set as 5, so five attentional regions are found. For EdgeBox, we directly use the codes provided by \cite{zitnick2014edge} to extract the proposals, and adopt non-maximum suppression (NMS) with a threshold of 0.7 on them based on their objectness scores to exclude the seriously overlapped proposals. We also visualize the top five ones for a fair comparison. Figure \ref{fig:attention_region} shows that the regions generated by our model better capture the discriminative regions (e.g., the head part of dogs), and most of them concentrate on the area of semantic objects. For EdgeBox, although its top-5 proposals cover most objects in the given image, most of them contain non-object areas that carry less discriminative information for classification. 

In order to clearly show the advantages of attentional region localization, we conduct experiments to compare the classification performance when using attentional regions or object proposals in the same framework. To this end, we first remove the spatial transformer and replace the attentional region with the selected five object proposals, with the other components left unchanged. Table \ref{table:voc_attention} gives the results on the VOC 2007 dataset. It is shown that attentional regions lead to better performance. In fact, proposal-based methods need hundreds of regions or even more proposals to cover most objects. Our model also achieves better performance than those using hundreds of proposals, such as HCP and FeV+LV, which use 500 and 400 proposals, respectively (see Table \ref{table:comparision_voc07}).

\begin{table}[htp]
\centering
\begin{tabular}{c|c}
\hline
\centering  type  &   mAP\\
\hline
\hline
object proposals  & 88.6 \\
attentional regions  & 90.4 \\
\hline
\end{tabular}
\vspace{4pt}
\caption{Comparison of the mAPs of our model using attentional regions and object proposals, respectively, on the PASCAL VOC 2007 dataset. The results are all evaluated using single-crop at scale of 512$\times$512}
\label{table:voc_attention}
\end{table}

\subsubsection{Analysis of the attentional region constraints}
We propose three types of novel constraints for attentional region localization, facilitating the task of multi-label image classification. To validate their contributions, we remove all three constraints, and retrain the model on the VOC 2007 dataset. The results, depicted in Table \ref{table:voc_constraints}, show a significant drop in mAP, well demonstrating the effectiveness of the constraints as a whole. We further remove one of three constraints, and retrain the model to evaluate the effectiveness of each constraint individually. The performance also declines when any constraint is excluded (see Table \ref{table:voc_constraints}). Therefore, it suggests that all of the three constraints are of importance for our model, they work cooperatively to facilitate the improvement of classification. We also conduct similar experiments on the MS-COCO dataset. As Table \ref{table:coco_constraints} shown, although MS-COCO is far different from VOC, similar results have been observed, again demonstrating their contributions on various scenes.

\begin{table}[htp]
\centering
\begin{tabular}{c|c}
\hline
\centering   constraints   & mAP\\
\hline
\hline
\centering    null    &   89.9   \\
\centering    S+P &     90.2   \\
\centering    A+S &     90.4   \\
\centering    A+P &     90.3   \\
\centering    A+S+P    &   91.3   \\
\hline
\end{tabular}
\vspace{4pt}
\caption{Comparison of mAP of our model learned using different constraints of attentional region localization on the PASCAL VOC 2007 dataset. The results are all evaluated using multi-crop at the scale of 512$\times$512. We abbreviate anchor, scale and positive constraints as A, S, P for simple illustration.}
\label{table:voc_constraints}
\end{table}

\begin{table}[htp]
\centering
\begin{tabular}{c|c|c|c}
\hline
\centering  constraints  &  C-F1 &  O-F1  & mAP\\
\hline
\hline
\centering     null & 65.8 & 70.9  &   71.5   \\
\centering    A+S+P & 66.5 &  71.3 &   72.2   \\
\hline
\end{tabular}
\vspace{4pt}
\caption{Comparison of C-F1, O-F1 and mAP of our model learned with and without constraints of attentional region localization on the MS-COCO dataset. The results are all evaluated using multi-crop at the scale of 512$\times$512. We abbreviate anchor, scale and positive constraints as A, S, P for simple illustration.}
\label{table:coco_constraints}
\end{table}

\begin{table}[htp]
\centering
\begin{tabular}{c|c|c}
\hline
\centering      & VOC 2007 & MS-COCO \\
\hline
\hline
\centering    s=512 + single-crop &     90.4   &     70.4 \\
\centering    s=640 + single-crop &     90.4  &     70.5\\
\centering    s=512 + ten-crop &     91.3  &     72.2 \\
\centering    s=640 + ten-crop &     91.3 &     72.3  \\
\centering    two scales + ten-crop &   91.9  &   73.4 \\
\hline
\end{tabular}
\vspace{4pt}
\caption{Comparison of mAP with multi-scale and multi-crop on the PASCAL VOC 2007 and MS-COCO datasets.}
\label{table:voc_msmc}
\end{table}

\subsubsection{Multi-scale multi-view evaluation}
We assess the impact of fusion of multi-scale and multi-crop at the test stage. Two scales ($512 \times 512$ and $640 \times 640$) are used in our experiments. For each scale, we extract ten crop features. Hence, we reported the performance of single-scale + single-crop, single-scale + multi-crop and multi-scale + multi-crop, in Table \ref{table:voc_msmc}.
The results show that aggregating information from multi-crop on a single-scale can boost the performance, and fusing the results of both of two scale shows a further improvement.

\section{Conclusion}
In this paper, we have introduced a recurrent memorized-attention module into the deep neural network architecture to solve the problem of multi-label image recognition. Specifically, our proposed recurrent memorized-attention module is composed of a spatial transformer layer for localizing attentional regions from the image and an LSTM unit to predict the labeling score based on the feature of a localized region and preserve the past information for the located regions. Experimental results on large-scale benchmarks (e.g., PASCAL VOC, COCO) demonstrate that our proposed deep model can significantly improve the state of the art in both accuracy and efficiency.

{\small
\bibliographystyle{ieee}
\bibliography{egbib}
}

\end{document}